%% file: main.tex
\documentclass[conference,letterpaper]{IEEEtran}
\IEEEoverridecommandlockouts
\usepackage[dvipsnames]{xcolor}
\usepackage{cite,capt-of}
\usepackage{xurl}
\usepackage{microtype}
\AtBeginDocument{\microtypesetup{activate=false}}
\usepackage{stfloats}
\providecommand{\captionsetup}[1]{}
\providecommand{\ie}{\emph{i.e.}}
\input{preamble}
\usepackage[hidelinks]{hyperref}
\usepackage[capitalize,nameinlink]{cleveref}
\title{\method{}: Beyond Sim(3)\\for Metric Upgrade of Feed-Forward Geometry\\with Visual-Inertial Anchors}

\author{%
  \IEEEauthorblockN{Lingxiang Hu\textsuperscript{1}, Zhenjun Zhao\textsuperscript{2}, Qi Shao\textsuperscript{3},
    Naima Ait Oufroukh\textsuperscript{1}, Fabien Bonardi\textsuperscript{1},\\
    Raymond Ghandour\textsuperscript{4}, Shenghai Yuan\textsuperscript{5},
    Javier Civera\textsuperscript{2}, Baoru Huang\textsuperscript{3}}
  \IEEEauthorblockA{\textsuperscript{1}Université Paris-Saclay, Univ Évry, IBISC, France \quad
    \textsuperscript{2}Universidad de Zaragoza, Spain \quad
    \textsuperscript{3}University of Liverpool, UK\\
    \textsuperscript{4}American University of the Middle East, Kuwait \quad
    \textsuperscript{5}Nanyang Technological University, Singapore}%
}
\newcommand{\method}{ProManO}
\IEEEaftertitletext{%
  \par\noindent\begin{minipage}{\textwidth}
  \centering
  \includegraphics[width=\textwidth]{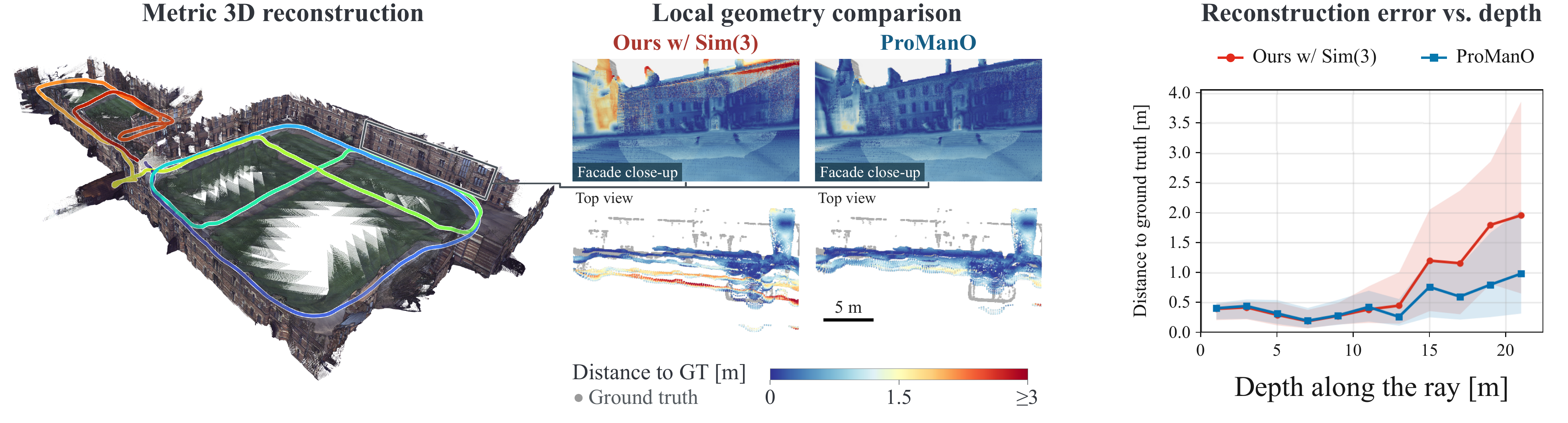}

  \captionof{figure}{Left and center: Reconstruction and depth-dependent errors, for Sim(3) and ProManO. Right: For identical DA3 pixels in one submap, lines show median nearest-neighbor distances to LiDAR ground truth in $2$\,m depth bins. Shading marks the 25th--75th percentiles.}
  \label{fig:teaser}

  \end{minipage}
  \par\vspace{10pt}%
}
\input{layout_floats}

\begin{document}
\maketitle
\input{abstract}

\input{introduction}
\input{related_work}
\LayoutPipeline
\input{residual_ambiguity}
\input{method.tex}

\input{experiments}
\input{conclusion}
\bibliographystyle{IEEEtran}
\bibliography{references}
\end{document}

%% file: preamble.tex
\usepackage[T1]{fontenc}
\usepackage{times,graphicx,amsmath,amssymb,amsthm,booktabs,multirow,tabularx}

\definecolor{mapblue}{RGB}{44,91,126}
\definecolor{mapgreen}{RGB}{52,120,93}

\theoremstyle{plain}
\newtheorem{theorem}{Theorem}
\newtheorem{lemma}{Lemma}
\newtheorem{proposition}{Proposition}

\theoremstyle{remark}

\AtBeginDocument{
\crefname{theorem}{Theorem}{Theorems}
\crefname{lemma}{Lemma}{Lemmas}
\crefname{proposition}{Proposition}{Propositions}
\crefname{corollary}{Corollary}{Corollaries}
\crefname{remark}{Remark}{Remarks}
}

\providecommand{\bestval}[1]{\textbf{#1}}
\providecommand{\secondval}[1]{\underline{#1}}
\providecommand{\thirdval}[1]{\textit{#1}}
\makeatletter
\g@addto@macro\normalsize{%
  \setlength\abovedisplayskip{4pt plus 2pt minus 2pt}%
  \setlength\belowdisplayskip{4pt plus 2pt minus 2pt}%
  \setlength\abovedisplayshortskip{0pt plus 2pt}%
  \setlength\belowdisplayshortskip{2pt plus 2pt minus 1pt}%
}
\makeatother
\newcommand{\PAR}[1]{\vspace{0.1cm}\noindent{\bf #1} }

%% file: layout_floats.tex
\newcommand{\LayoutPipeline}{%
\begin{figure*}[tp]
\centering
\includegraphics[width=\textwidth]{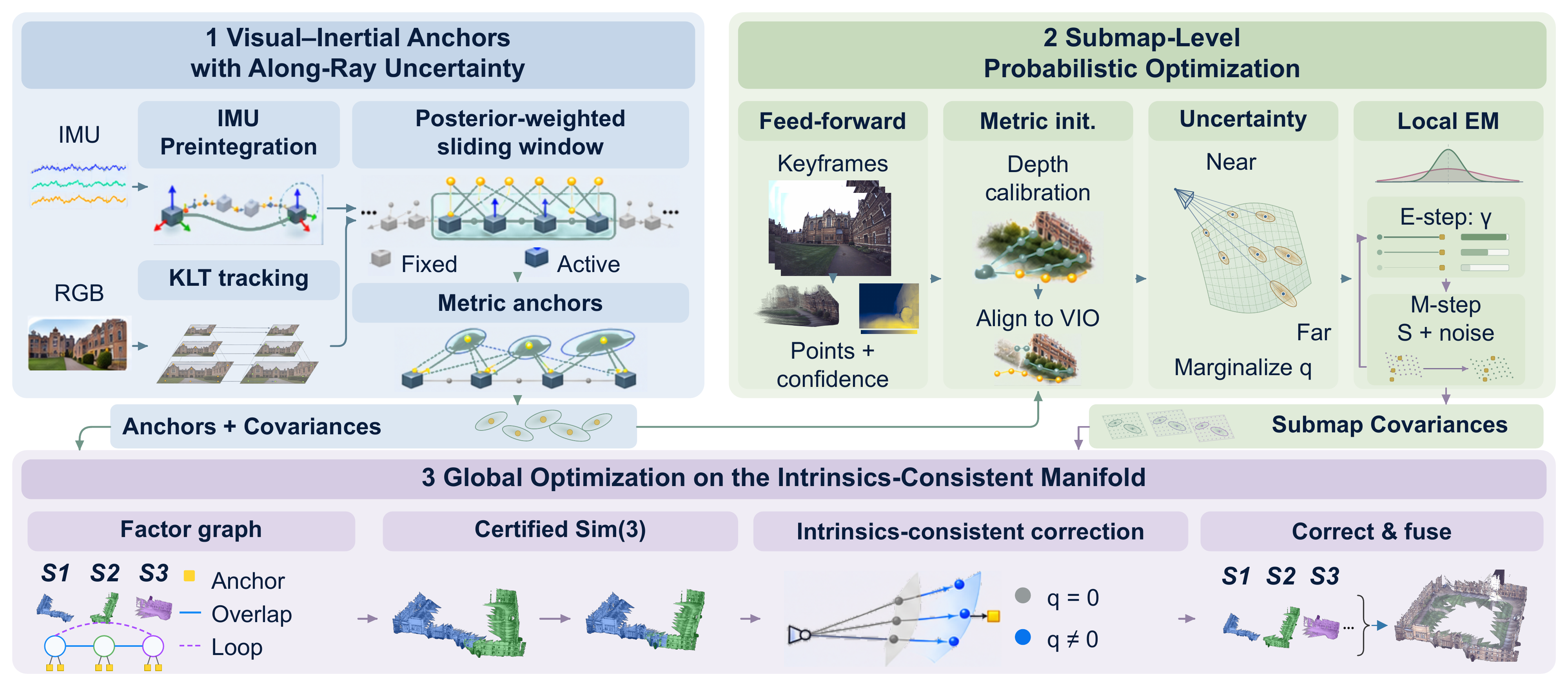}
\vspace{-5mm}
\caption{Overview of \method{}. The proposed framework includes main stages: visual-inertial frontend (\cref{sec:vi_anchors}, upper left), submap-level probabilistic optimization (\cref{sec:submap_alignment}, upper right), and global intrinsics-consistent optimization (\cref{sec:global_alignment}, bottom).}
\label{fig:pipeline}

\end{figure*}
}

\newcommand{\LayoutQualitative}{%
\begin{figure*}[tp]
\centering
\includegraphics[width=0.96\linewidth]{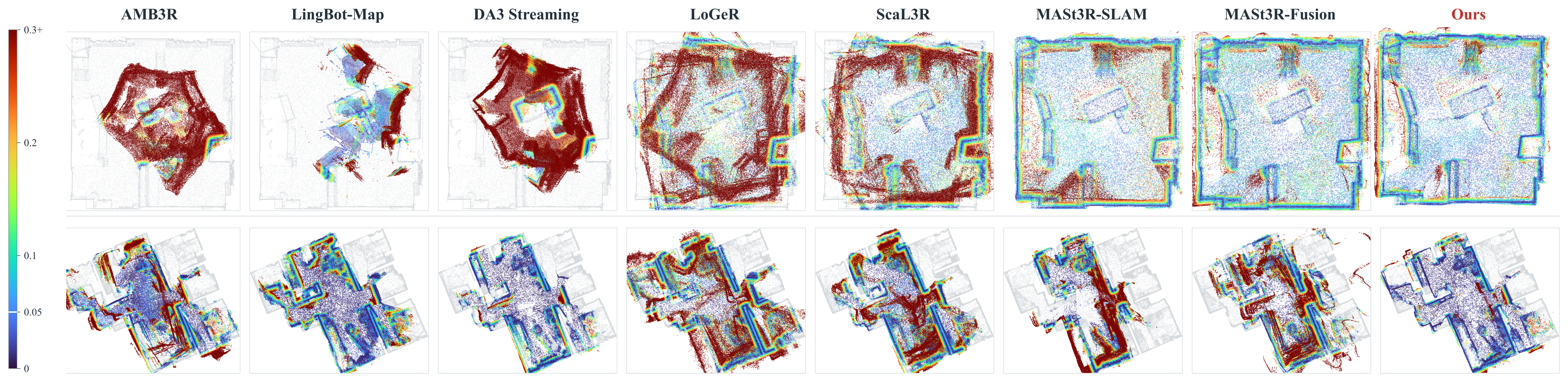}
\caption{Reconstruction results of \method{} and baselines on EuRoC V1\_03 (top) and CID-SIMS apartment1\_1 (bottom) sequences.}
\label{fig:qualitative_reconstruction}

\end{figure*}
}

\newcommand{\LayoutCloseups}{%
\begin{figure*}[tp]
\centering
\includegraphics[width=\linewidth]{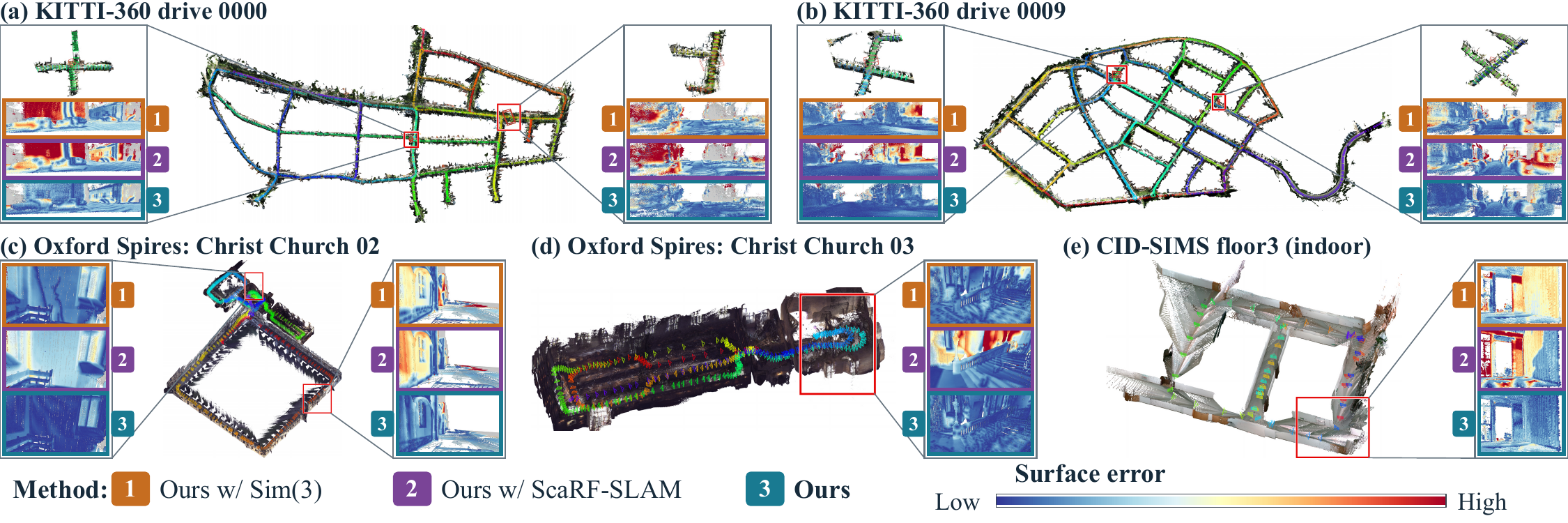}%
\vspace{-2mm}%
\caption{Qualitative reconstruction of \method{} and two variants on outdoor and indoor sequences.}
\label{fig:region_closeups}

\end{figure*}
}

\newcommand{\LayoutSequenceMaps}{%
\begin{figure*}[tp]
\centering
\includegraphics[width=\textwidth]{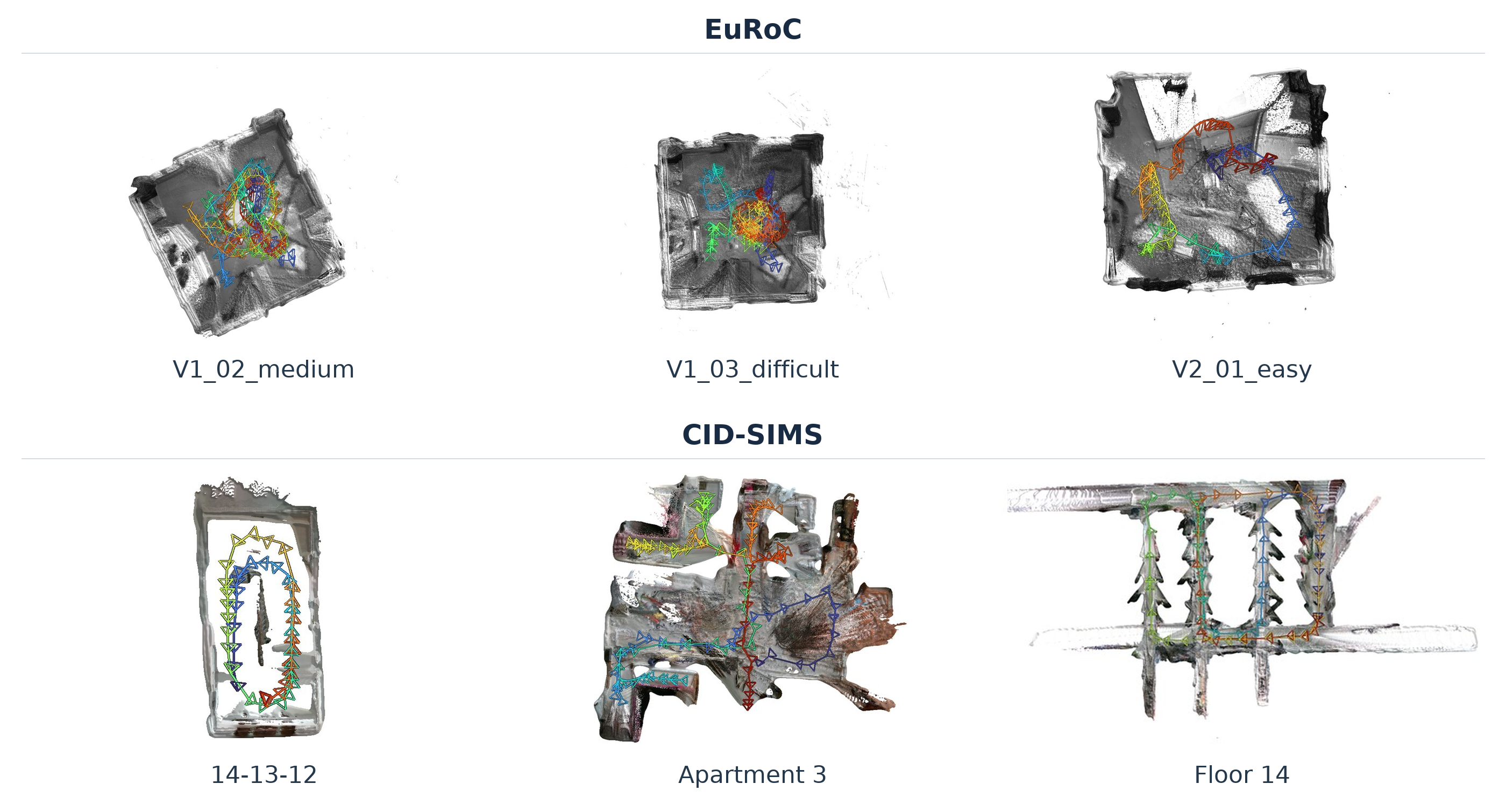}
\caption{Full-sequence indoor reconstructions on EuRoC (top) and CID-SIMS (bottom). Colored camera poses trace the estimated trajectory through each point cloud.}
\label{fig:indoor_sequence_maps}
\end{figure*}
}

\newcommand{\LayoutOutdoorSequenceMaps}{%
\begin{figure*}[tp]
\centering
\includegraphics[width=\textwidth]{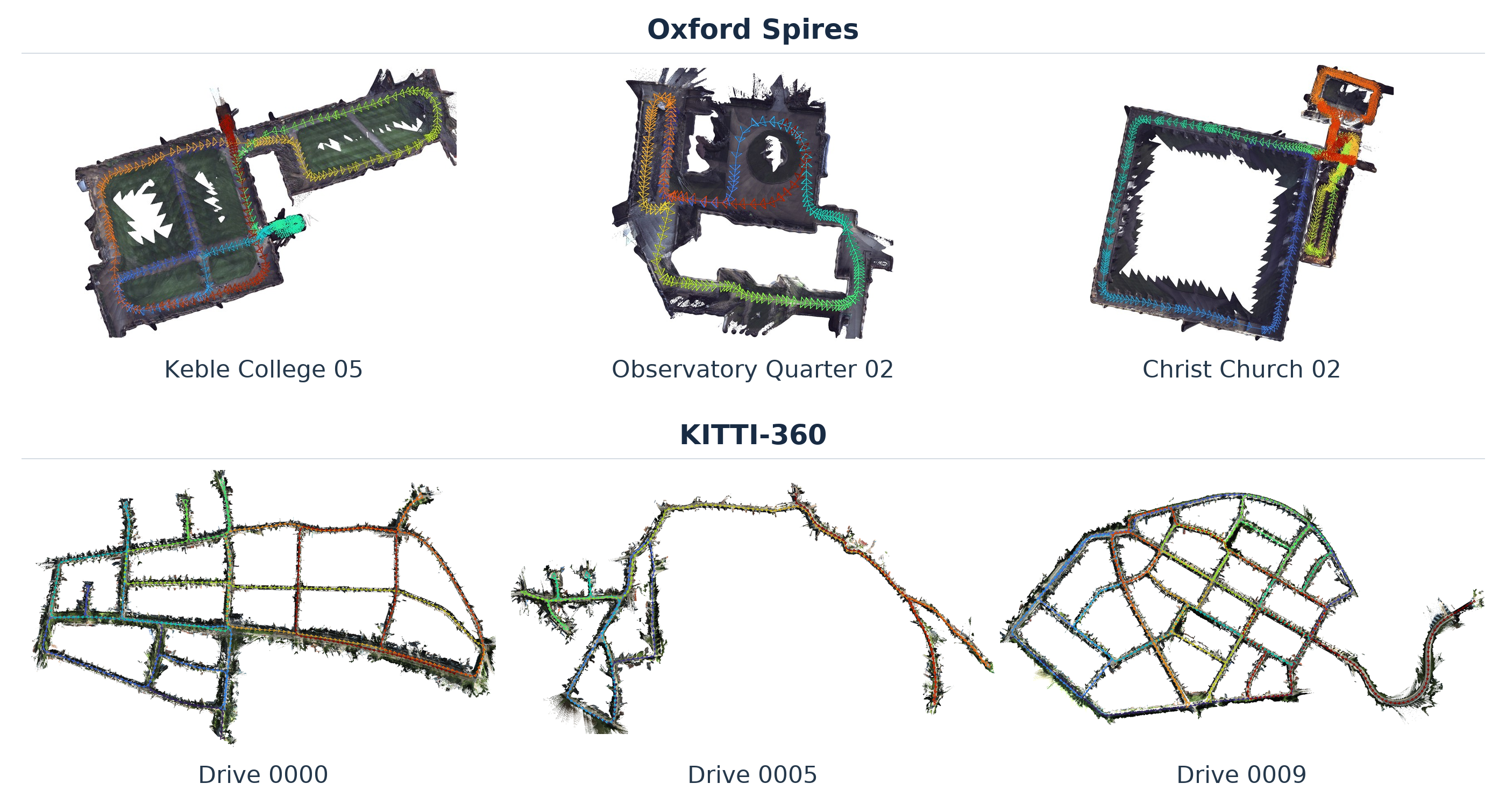}
\caption{Full-sequence outdoor reconstructions on Oxford Spires (top) and KITTI-360 (bottom). The top-down views show the extent of each reconstructed site and the estimated camera trajectory.}
\label{fig:outdoor_sequence_maps}
\end{figure*}
}

\input{tables/tab_recon_main}

%% file: tables/tab_recon_main.tex
\newcommand{\LayoutIndoorAte}{%
\begin{table*}[tp]
\centering
\captionsetup{font=small}
\caption{Indoor localization (ATE RMSE [m] $\downarrow$) on EuRoC~\cite{burri2016euroc} and CID-SIMS~\cite{zhang2024cidsims}. $\times$: tracking failure.}
\label{tab:ate_indoor}
\vspace{-2mm}
\scriptsize
\setlength{\tabcolsep}{1.7pt}
\renewcommand{\arraystretch}{1.12}
\begin{tabular*}{\linewidth}{@{\extracolsep{\fill}}lrrrrrrrrrrrrrrrr@{}}
\toprule
 & \multicolumn{7}{c}{\textbf{EuRoC}} & \multicolumn{9}{c}{\textbf{CID-SIMS}} \\
\cmidrule(lr){2-8} \cmidrule(lr){9-17}
Method & V101 & V102 & V103 & V201 & V202 & V203 & Avg. & 14-13-12 & 14-13-14 & A1-1 & A2-1 & A3-1 & F13-1 & F14-1 & F3-1 & Avg. \\
\midrule
LingBot-Map~\cite{chen2026lingbot} & 0.451 & 0.686 & 0.683 & 0.552 & 0.555 & 0.758 & 0.614 & 1.375 & 7.840 & 0.211 & 0.188 & 0.241 & 11.481 & 1.163 & 2.703 & 3.150 \\
LoGeR~\cite{zhang2026loger} & 0.569 & 0.583 & 0.580 & 1.258 & 0.655 & 0.624 & 0.711 & 0.342 & 9.558 & 0.431 & 0.755 & 0.266 & 9.049 & 1.365 & 1.895 & 2.958 \\
AMB3R~\cite{wang2026amb3r} & 0.491 & 1.513 & 1.219 & 0.456 & 1.190 & 1.682 & 1.092 & 0.404 & 5.986 & 0.321 & 2.300 & 2.768 & 6.271 & 0.655 & 0.449 & 2.394 \\
MASt3R-SLAM~\cite{murai2025mast3rslam} & 0.127 & 0.085 & 0.132 & 0.157 & 0.107 & 0.152 & 0.127 & 1.945 & $\times$ & 1.539 & $\times$ & 2.465 & 2.705 & 4.287 & \thirdval{0.274} & $\times$ \\
DA3 Streaming~\cite{lin2026depthanything3} & 0.766 & 1.351 & 1.019 & 1.103 & 0.704 & 1.229 & 1.029 & 1.600 & 1.900 & 0.259 & 0.503 & \secondval{0.117} & 1.874 & \bestval{0.123} & \bestval{0.122} & 0.812 \\
\midrule
ORB-SLAM3~\cite{campos2021orbslam3} & \bestval{0.033} & \bestval{0.017} & \secondval{0.038} & \thirdval{0.060} & \bestval{0.015} & \bestval{0.018} & \bestval{0.030} & $\times$ & $\times$ & $\times$ & 0.184 & $\times$ & $\times$ & 0.783 & $\times$ & $\times$ \\
OpenVINS~\cite{geneva2020openvins} & 0.069 & 0.058 & \thirdval{0.066} & 0.104 & 0.067 & 0.178 & 0.090 & 0.139 & 2.012 & $\times$ & \thirdval{0.082} & $\times$ & 2.348 & 0.410 & $\times$ & $\times$ \\
VINS-Mono~\cite{qin2018vinsmono} & 0.075 & 0.098 & 0.189 & 0.085 & 0.107 & 0.289 & 0.141 & 0.178 & 2.060 & 0.180 & \secondval{0.075} & \bestval{0.095} & 1.204 & \thirdval{0.310} & 0.785 & \thirdval{0.611} \\
VINS-Fusion~\cite{qin2019vinsfusion} & 0.059 & 0.159 & 0.160 & $\times$ & 0.119 & 0.198 & $\times$ & \secondval{0.100} & \secondval{0.914} & \thirdval{0.135} & 0.087 & \thirdval{0.133} & 0.897 & 0.581 & 1.881 & \secondval{0.591} \\
Kimera-VIO~\cite{rosinol2020kimera} & 0.163 & 0.092 & 0.172 & 0.113 & 0.223 & 0.626 & 0.231 & 0.293 & $\times$ & 0.784 & 0.242 & 0.233 & 4.660 & 2.315 & 3.644 & $\times$ \\
OKVIS2~\cite{leutenegger2022okvis2} & 0.106 & 0.080 & 0.087 & 0.133 & 0.104 & 0.173 & 0.114 & \thirdval{0.128} & \thirdval{1.316} & 0.291 & 0.377 & 0.280 & 1.060 & 0.908 & 0.748 & 0.638 \\
DM-VIO~\cite{stumberg2022dmvio} & \thirdval{0.057} & \thirdval{0.040} & 0.096 & \bestval{0.028} & \secondval{0.040} & \thirdval{0.101} & \thirdval{0.060} & \bestval{0.099} & 1.486 & \secondval{0.133} & 0.166 & 0.245 & \secondval{0.452} & 0.417 & 5.385 & 1.048 \\
MASt3R-Fusion~\cite{zhou2026mast3rfusion} & 0.065 & \secondval{0.034} & \bestval{0.023} & 0.107 & \thirdval{0.041} & \secondval{0.032} & \secondval{0.050} & 0.210 & 3.852 & 0.578 & 0.148 & 0.263 & \bestval{0.343} & 0.884 & 1.459 & 0.967 \\
\midrule
\textbf{Ours} & \secondval{0.046} & 0.094 & 0.099 & \secondval{0.055} & 0.091 & 0.118 & 0.084 & 0.180 & \bestval{0.717} & \bestval{0.065} & \bestval{0.065} & 0.181 & \thirdval{0.610} & \secondval{0.182} & \secondval{0.143} & \bestval{0.268} \\
\bottomrule
\end{tabular*}
\end{table*}
}

\newcommand{\LayoutOutdoorAte}{%
\begin{table*}[tp]
\centering
\captionsetup{font=small}
\caption{Outdoor localization (ATE RMSE [m] $\downarrow$) on Oxford Spires~\cite{tao2026oxfordspires} and KITTI-360~\cite{liao2022kitti360}. $\times$: tracking failure.}
\label{tab:ate_outdoor}
\vspace{-2mm}
\scriptsize
\setlength{\tabcolsep}{1.7pt}
\renewcommand{\arraystretch}{1.12}
\begin{tabular*}{\linewidth}{@{\extracolsep{\fill}}lrrrrrrrrrrrrrrrr@{}}
\toprule
 & \multicolumn{10}{c}{\textbf{Oxford Spires}} & \multicolumn{6}{c}{\textbf{KITTI-360}} \\
\cmidrule(lr){2-11} \cmidrule(lr){12-17}
Method & K-02 & K-04 & K-05 & O-01 & O-02 & BP-01 & BP-05 & CC-02 & CC-03 & Avg. & 00 & 02 & 05 & 06 & 09 & Avg. \\
\midrule
LingBot-Map~\cite{chen2026lingbot} & 25.75 & 39.99 & 34.11 & 26.26 & 25.00 & 6.65 & 43.19 & 37.23 & 2.80 & 26.78 & 320.14 & 389.90 & 382.87 & 306.87 & 255.44 & 331.04 \\
LoGeR~\cite{zhang2026loger} & 6.09 & 5.39 & 7.10 & 8.23 & 9.98 & 10.80 & 5.59 & 7.48 & 2.36 & 7.00 & 281.38 & $\times$ & 159.05 & 278.61 & $\times$ & $\times$ \\
AMB3R~\cite{wang2026amb3r} & 12.41 & 3.55 & 41.15 & 2.35 & 3.11 & 50.95 & 10.38 & 42.98 & 1.11 & 18.67 & 308.77 & 257.63 & 349.33 & 280.80 & 238.08 & 286.92 \\
MASt3R-SLAM~\cite{murai2025mast3rslam} & 3.62 & $\times$ & 12.21 & 2.34 & 1.78 & $\times$ & $\times$ & $\times$ & 0.56 & $\times$ & $\times$ & $\times$ & $\times$ & $\times$ & $\times$ & $\times$ \\
DA3 Streaming~\cite{lin2026depthanything3} & 2.03 & 10.61 & 2.67 & 2.28 & 1.36 & 8.96 & 7.42 & 8.80 & 2.20 & 5.15 & 253.68 & 303.13 & 339.31 & 238.97 & 238.28 & 274.67 \\
\midrule
ORB-SLAM3~\cite{campos2021orbslam3} & \thirdval{0.48} & \bestval{0.36} & \bestval{0.30} & \secondval{0.38} & \secondval{0.35} & 1.76 & \thirdval{0.56} & \thirdval{0.68} & \secondval{0.16} & \secondval{0.56} & $\times$ & $\times$ & 24.62 & $\times$ & $\times$ & $\times$ \\
OpenVINS~\cite{geneva2020openvins} & 0.62 & 1.06 & \thirdval{0.98} & 0.69 & 0.66 & \thirdval{0.71} & $\times$ & 1.25 & 0.42 & $\times$ & 16.47 & $\times$ & $\times$ & $\times$ & $\times$ & $\times$ \\
VINS-Mono~\cite{qin2018vinsmono} & \secondval{0.42} & 1.62 & 1.33 & 1.32 & \thirdval{0.55} & \bestval{0.53} & \bestval{0.37} & 1.35 & \thirdval{0.29} & \thirdval{0.86} & 27.08 & 288.76 & \thirdval{7.77} & 15.98 & $\times$ & $\times$ \\
VINS-Fusion~\cite{qin2019vinsfusion} & 2.09 & $\times$ & $\times$ & $\times$ & $\times$ & 0.96 & $\times$ & $\times$ & $\times$ & $\times$ & 11.49 & $\times$ & $\times$ & 12.61 & $\times$ & $\times$ \\
Kimera-VIO~\cite{rosinol2020kimera} & 5.81 & $\times$ & $\times$ & 7.82 & $\times$ & $\times$ & 7.11 & 5.34 & $\times$ & $\times$ & $\times$ & $\times$ & $\times$ & $\times$ & $\times$ & $\times$ \\
OKVIS2~\cite{leutenegger2022okvis2} & 0.96 & 1.64 & 1.60 & 1.84 & 0.88 & 1.20 & 0.75 & 1.68 & 0.51 & 1.23 & $\times$ & $\times$ & $\times$ & $\times$ & $\times$ & $\times$ \\
DM-VIO~\cite{stumberg2022dmvio} & 3.84 & \thirdval{0.56} & $\times$ & \thirdval{0.39} & $\times$ & \secondval{0.59} & 0.84 & \secondval{0.54} & 0.37 & $\times$ & \thirdval{10.07} & \thirdval{11.29} & 9.68 & \thirdval{7.51} & \secondval{9.35} & \thirdval{9.58} \\
MASt3R-Fusion~\cite{zhou2026mast3rfusion} & 2.28 & 1.96 & 1.21 & 0.72 & 0.75 & 1.46 & 4.59 & 0.76 & 0.32 & 1.56 & \secondval{2.89} & \secondval{9.65} & \secondval{6.50} & \bestval{3.02} & \thirdval{22.75} & \secondval{8.96} \\
\midrule
\textbf{Ours} & \bestval{0.23} & \secondval{0.49} & \secondval{0.40} & \bestval{0.32} & \bestval{0.32} & 0.79 & \secondval{0.51} & \bestval{0.29} & \bestval{0.13} & \bestval{0.39} & \bestval{2.75} & \bestval{9.17} & \bestval{2.29} & \secondval{4.99} & \bestval{3.71} & \bestval{4.58} \\
\bottomrule
\end{tabular*}
\end{table*}
}

\newcommand{\LayoutEurocRecon}{%
\begin{table*}[tp]
\centering
\captionsetup{font=small}
\caption{Reconstruction on EuRoC~\cite{burri2016euroc} (CD [m] $\downarrow$ and $F_1$ [\%] $\uparrow$ at $5$\,cm).}
\label{tab:recon_euroc}
\vspace{-2mm}
\footnotesize
\setlength{\tabcolsep}{1.6pt}
\renewcommand{\arraystretch}{1.12}
\begin{tabular*}{\linewidth}{@{\extracolsep{\fill}}lrrrrrrrrrrrrrr@{}}
\toprule
\multirow{2}{*}{Method} & \multicolumn{2}{c}{V101} & \multicolumn{2}{c}{V102} & \multicolumn{2}{c}{V103} & \multicolumn{2}{c}{V201} & \multicolumn{2}{c}{V202} & \multicolumn{2}{c}{V203} & \multicolumn{2}{c}{Avg.} \\
\cmidrule(lr){2-3} \cmidrule(lr){4-5} \cmidrule(lr){6-7} \cmidrule(lr){8-9} \cmidrule(lr){10-11} \cmidrule(lr){12-13} \cmidrule(lr){14-15}
 & CD & $F_1$ & CD & $F_1$ & CD & $F_1$ & CD & $F_1$ & CD & $F_1$ & CD & $F_1$ & CD & $F_1$ \\
\midrule
LingBot-Map~\cite{chen2026lingbot} & 1.091 & 8.01 & 1.887 & 0.19 & 1.118 & 6.26 & 0.783 & 13.96 & 1.094 & 1.25 & 0.699 & 13.88 & 1.112 & 7.26 \\
LoGeR~\cite{zhang2026loger} & 0.837 & 16.20 & 0.685 & 16.90 & 0.532 & 16.31 & 0.801 & 8.86 & 0.592 & 13.21 & 0.569 & 7.85 & 0.669 & 13.22 \\
ScaL3R~\cite{xie2026scal3r} & 0.850 & 14.91 & 0.701 & 16.41 & 0.460 & 22.09 & 0.749 & 15.74 & 0.568 & 18.73 & 0.564 & 13.65 & 0.648 & 16.92 \\
AMB3R~\cite{wang2026amb3r} & 0.937 & 13.42 & 1.489 & 2.89 & 1.200 & 1.81 & 0.904 & 5.91 & 0.789 & 6.18 & 1.535 & 0.50 & 1.142 & 5.12 \\
MASt3R-SLAM~\cite{murai2025mast3rslam} & 0.493 & 43.45 & 0.305 & \secondval{56.83} & 0.268 & \thirdval{47.52} & 0.427 & 43.47 & 0.252 & \bestval{49.15} & 0.268 & \bestval{48.80} & 0.336 & \thirdval{48.20} \\
DA3 Streaming~\cite{lin2026depthanything3} & 1.022 & 12.09 & 1.260 & 1.26 & 1.065 & 1.95 & 0.836 & 13.04 & 0.642 & 15.86 & 0.978 & 4.12 & 0.967 & 8.05 \\
\midrule
MASt3R-Fusion~\cite{zhou2026mast3rfusion} & 0.453 & 41.35 & 0.289 & 50.32 & 0.248 & \secondval{48.38} & 0.448 & 28.78 & 0.299 & 40.68 & 0.296 & \thirdval{43.52} & 0.339 & 42.17 \\
\midrule
Ours w/ Sim(3) & \secondval{0.064} & \secondval{59.18} & \secondval{0.056} & \thirdval{56.16} & \secondval{0.070} & 45.33 & \secondval{0.081} & \thirdval{54.86} & \secondval{0.089} & 40.58 & \secondval{0.075} & 43.10 & \secondval{0.073} & \secondval{49.87} \\
Ours w/ ScaRF-SLAM~\cite{zhang2026scarfs} & \thirdval{0.127} & \thirdval{51.61} & \thirdval{0.072} & 55.81 & \thirdval{0.110} & 38.84 & \thirdval{0.102} & \secondval{59.51} & \thirdval{0.126} & \secondval{48.10} & \thirdval{0.115} & 31.38 & \thirdval{0.109} & 47.54 \\
\textbf{Ours} & \bestval{0.054} & \bestval{64.80} & \bestval{0.053} & \bestval{56.96} & \bestval{0.063} & \bestval{50.13} & \bestval{0.065} & \bestval{61.96} & \bestval{0.084} & \thirdval{43.88} & \bestval{0.069} & \secondval{46.19} & \bestval{0.065} & \bestval{53.99} \\
\bottomrule
\end{tabular*}
\end{table*}
}

\newcommand{\LayoutCidRecon}{%
\begin{table*}[tp]
\centering
\captionsetup{font=small}
\caption{Reconstruction on CID-SIMS~\cite{zhang2024cidsims} (CD [m] $\downarrow$ and $F_1$ [\%] $\uparrow$ at $5$\,cm).}
\label{tab:recon_cid}
\vspace{-2mm}
\footnotesize
\setlength{\tabcolsep}{1.6pt}
\renewcommand{\arraystretch}{1.12}
\begin{tabular*}{\linewidth}{@{\extracolsep{\fill}}lrrrrrrrrrrrrrrrrrr@{}}
\toprule
\multirow{2}{*}{Method} & \multicolumn{2}{c}{14-13-12} & \multicolumn{2}{c}{14-13-14} & \multicolumn{2}{c}{A1-1} & \multicolumn{2}{c}{A2-1} & \multicolumn{2}{c}{A3-1} & \multicolumn{2}{c}{F13-1} & \multicolumn{2}{c}{F14-1} & \multicolumn{2}{c}{F3-1} & \multicolumn{2}{c}{Avg.} \\
\cmidrule(lr){2-3} \cmidrule(lr){4-5} \cmidrule(lr){6-7} \cmidrule(lr){8-9} \cmidrule(lr){10-11} \cmidrule(lr){12-13} \cmidrule(lr){14-15} \cmidrule(lr){16-17} \cmidrule(lr){18-19}
 & CD & $F_1$ & CD & $F_1$ & CD & $F_1$ & CD & $F_1$ & CD & $F_1$ & CD & $F_1$ & CD & $F_1$ & CD & $F_1$ & CD & $F_1$ \\
\midrule
LingBot-Map~\cite{chen2026lingbot} & 0.239 & 21.71 & 1.425 & 0.97 & 0.263 & 39.64 & \secondval{0.122} & \secondval{50.77} & 0.183 & 43.79 & 2.658 & 6.21 & 0.265 & 29.92 & 0.546 & 23.65 & 0.712 & 27.08 \\
LoGeR~\cite{zhang2026loger} & 0.182 & 32.58 & 1.027 & 9.45 & \secondval{0.203} & 39.71 & 0.160 & 40.63 & \secondval{0.139} & \thirdval{46.44} & 2.220 & 5.51 & 0.464 & 26.96 & 0.540 & 13.68 & 0.617 & 26.87 \\
ScaL3R~\cite{xie2026scal3r} & 0.324 & 18.25 & 1.197 & 4.28 & 0.258 & 40.66 & 0.329 & 24.90 & 0.306 & 22.05 & 1.281 & 4.83 & 0.318 & 22.13 & 0.482 & 16.72 & 0.562 & 19.23 \\
AMB3R~\cite{wang2026amb3r} & 0.206 & 20.51 & 0.719 & 4.52 & 0.235 & 28.75 & 0.286 & 24.49 & 0.271 & 22.13 & 1.124 & 10.29 & 0.422 & 22.37 & 0.477 & 21.64 & 0.467 & 19.34 \\
MASt3R-SLAM~\cite{murai2025mast3rslam} & 0.332 & 26.09 & $\times$ & $\times$ & 0.361 & 27.64 & $\times$ & $\times$ & 0.410 & 14.55 & 0.790 & 9.18 & 0.861 & 7.56 & 0.449 & 32.65 & $\times$ & $\times$ \\
DA3 Streaming~\cite{lin2026depthanything3} & 0.313 & 13.68 & 0.692 & 6.27 & 0.248 & 44.41 & 0.210 & 37.52 & 0.195 & 38.95 & 0.394 & 21.97 & \thirdval{0.224} & \thirdval{42.58} & 0.265 & \secondval{41.03} & 0.318 & 30.80 \\
\midrule
MASt3R-Fusion~\cite{zhou2026mast3rfusion} & \thirdval{0.160} & \secondval{45.06} & 0.603 & 11.22 & 0.227 & 34.83 & 0.211 & 39.33 & 0.303 & 23.96 & \thirdval{0.262} & \secondval{44.73} & 0.394 & 30.01 & 0.723 & 9.93 & 0.360 & 29.89 \\
\midrule
Ours w/ Sim(3) & 0.163 & \thirdval{34.25} & \bestval{0.300} & \bestval{24.46} & 0.228 & \thirdval{46.77} & \thirdval{0.130} & \thirdval{50.59} & \bestval{0.138} & \secondval{46.77} & \secondval{0.256} & \thirdval{38.61} & \secondval{0.220} & \secondval{45.49} & \secondval{0.156} & \thirdval{37.01} & \secondval{0.199} & \secondval{40.49} \\
Ours w/ ScaRF-SLAM~\cite{zhang2026scarfs} & \secondval{0.142} & 32.32 & \thirdval{0.476} & \thirdval{15.14} & \thirdval{0.206} & \bestval{56.79} & 0.269 & 44.32 & 0.331 & 27.76 & 0.294 & 36.37 & 0.379 & 37.13 & \thirdval{0.206} & 29.06 & \thirdval{0.288} & \thirdval{34.86} \\
\textbf{Ours} & \bestval{0.109} & \bestval{47.41} & \secondval{0.303} & \secondval{21.91} & \bestval{0.171} & \secondval{54.87} & \bestval{0.109} & \bestval{55.01} & \thirdval{0.140} & \bestval{47.37} & \bestval{0.248} & \bestval{44.92} & \bestval{0.220} & \bestval{46.18} & \bestval{0.134} & \bestval{44.26} & \bestval{0.179} & \bestval{45.24} \\
\bottomrule
\end{tabular*}
\end{table*}
}

\newcommand{\LayoutOxfordRecon}{%
\begin{table*}[tp]
\centering
\captionsetup{font=small}
\caption{Reconstruction on Oxford Spires~\cite{tao2026oxfordspires} (CD [m] $\downarrow$ and $F_1$ [\%] $\uparrow$ at $20$\,cm).}
\label{tab:recon_oxford}
\vspace{-2mm}
\scriptsize
\setlength{\tabcolsep}{1.1pt}
\renewcommand{\arraystretch}{1.12}
\begin{tabular*}{\linewidth}{@{\extracolsep{\fill}}lrrrrrrrrrrrrrrrrrrrr@{}}
\toprule
\multirow{2}{*}{Method} & \multicolumn{2}{c}{K-02} & \multicolumn{2}{c}{K-04} & \multicolumn{2}{c}{K-05} & \multicolumn{2}{c}{O-01} & \multicolumn{2}{c}{O-02} & \multicolumn{2}{c}{BP-01} & \multicolumn{2}{c}{BP-05} & \multicolumn{2}{c}{CC-02} & \multicolumn{2}{c}{CC-03} & \multicolumn{2}{c}{Avg.} \\
\cmidrule(lr){2-3} \cmidrule(lr){4-5} \cmidrule(lr){6-7} \cmidrule(lr){8-9} \cmidrule(lr){10-11} \cmidrule(lr){12-13} \cmidrule(lr){14-15} \cmidrule(lr){16-17} \cmidrule(lr){18-19} \cmidrule(lr){20-21}
 & CD & $F_1$ & CD & $F_1$ & CD & $F_1$ & CD & $F_1$ & CD & $F_1$ & CD & $F_1$ & CD & $F_1$ & CD & $F_1$ & CD & $F_1$ & CD & $F_1$ \\
\midrule
LingBot-Map~\cite{chen2026lingbot} & 6.71 & 5.82 & 11.08 & 4.04 & 7.48 & 5.79 & 8.61 & 2.63 & 8.16 & 3.10 & 2.11 & 16.78 & 8.09 & 2.79 & 9.07 & 5.60 & 2.64 & 32.97 & 7.11 & 8.84 \\
LoGeR~\cite{zhang2026loger} & 1.39 & 12.22 & 1.13 & 16.59 & 1.12 & 25.45 & 1.32 & 15.70 & 1.36 & 20.10 & 1.61 & 14.98 & 1.41 & 17.69 & 1.51 & 13.47 & 1.99 & 43.13 & 1.43 & 19.93 \\
AMB3R~\cite{wang2026amb3r} & 2.57 & 6.65 & 0.74 & 26.17 & 11.74 & 2.37 & \thirdval{0.85} & 27.59 & 0.87 & 24.42 & 9.99 & 3.51 & 1.04 & 29.13 & 10.34 & 2.06 & 2.11 & 40.05 & 4.47 & 17.99 \\
MASt3R-SLAM~\cite{murai2025mast3rslam} & 1.47 & 15.88 & $\times$ & $\times$ & 4.65 & 10.09 & 1.40 & 31.46 & 1.48 & 30.84 & $\times$ & $\times$ & $\times$ & $\times$ & $\times$ & $\times$ & 2.11 & 43.49 & $\times$ & $\times$ \\
DA3 Streaming~\cite{lin2026depthanything3} & \thirdval{1.01} & 22.81 & 3.33 & 11.00 & 0.92 & 26.84 & 1.94 & 24.12 & 1.76 & 35.05 & 2.31 & 10.99 & 2.62 & 11.51 & 3.61 & 5.19 & 2.37 & 36.92 & 2.21 & 20.49 \\
\midrule
MASt3R-Fusion~\cite{zhou2026mast3rfusion} & 1.73 & 24.55 & 1.25 & 34.24 & 1.13 & 37.46 & 1.50 & 40.50 & 1.51 & 41.67 & 1.60 & \secondval{41.75} & 1.82 & 22.15 & 0.97 & 42.90 & 1.87 & 50.89 & 1.49 & 37.35 \\
\midrule
Ours w/ Sim(3) & \secondval{0.71} & \thirdval{50.43} & \thirdval{0.45} & \thirdval{49.19} & \thirdval{0.45} & \thirdval{55.67} & \secondval{0.35} & \thirdval{45.49} & \thirdval{0.32} & \thirdval{53.23} & \thirdval{0.66} & 36.14 & \thirdval{0.74} & \thirdval{46.07} & \secondval{0.33} & \secondval{60.74} & \secondval{0.61} & \secondval{61.74} & \secondval{0.51} & \thirdval{50.97} \\
Ours w/ ScaRF-SLAM~\cite{zhang2026scarfs} & 1.14 & \secondval{54.55} & \secondval{0.43} & \secondval{50.74} & \secondval{0.35} & \secondval{59.42} & \secondval{0.35} & \secondval{50.70} & \secondval{0.30} & \secondval{57.88} & \secondval{0.64} & \thirdval{38.06} & \secondval{0.66} & \secondval{51.52} & \thirdval{0.46} & \thirdval{57.24} & \thirdval{0.65} & \thirdval{61.29} & \thirdval{0.55} & \secondval{53.49} \\
\textbf{Ours} & \bestval{0.48} & \bestval{61.15} & \bestval{0.33} & \bestval{57.41} & \bestval{0.30} & \bestval{66.18} & \bestval{0.27} & \bestval{59.66} & \bestval{0.25} & \bestval{64.53} & \bestval{0.45} & \bestval{44.97} & \bestval{0.42} & \bestval{59.64} & \bestval{0.28} & \bestval{67.00} & \bestval{0.55} & \bestval{66.12} & \bestval{0.37} & \bestval{60.74} \\
\bottomrule
\end{tabular*}
\end{table*}
}

\newcommand{\LayoutKittiRecon}{%
\begin{table}[tp]
\centering
\captionsetup{font=small}
\caption{Reconstruction on KITTI-360~\cite{liao2022kitti360} (CD [m] $\downarrow$ and $F_1$ [\%] $\uparrow$ at $20$\,cm).}
\label{tab:recon_kitti}
\vspace{-2mm}
\scriptsize
\setlength{\tabcolsep}{1.1pt}
\renewcommand{\arraystretch}{1.12}
\begin{tabular*}{\linewidth}{@{\extracolsep{\fill}}llrrrrrr@{}}
\toprule
Method & Metric & 00 & 02 & 05 & 06 & 09 & Avg. \\
\midrule
\multirow{2}{*}{LingBot-Map~\cite{chen2026lingbot}} & CD & 94.34 & 126.60 & 225.90 & 103.73 & 68.79 & 123.87 \\
 & $F_1$ & 0.21 & 0.00 & 0.00 & 0.00 & 0.37 & 0.12 \\
\multirow{2}{*}{LoGeR~\cite{zhang2026loger}} & CD & 80.03 & $\times$ & 28.99 & 76.02 & $\times$ & $\times$ \\
 & $F_1$ & 0.13 & $\times$ & 0.49 & 0.07 & $\times$ & $\times$ \\
\multirow{2}{*}{AMB3R~\cite{wang2026amb3r}} & CD & 33.11 & 25.37 & 87.93 & 91.61 & 39.79 & 55.56 \\
 & $F_1$ & 0.61 & 0.54 & 0.05 & 0.11 & 0.56 & 0.37 \\
\multirow{2}{*}{MASt3R-SLAM~\cite{murai2025mast3rslam}} & CD & $\times$ & $\times$ & $\times$ & $\times$ & $\times$ & $\times$ \\
 & $F_1$ & $\times$ & $\times$ & $\times$ & $\times$ & $\times$ & $\times$ \\
\multirow{2}{*}{DA3 Streaming~\cite{lin2026depthanything3}} & CD & 95.90 & 111.40 & 176.02 & 70.79 & 96.47 & 110.12 \\
 & $F_1$ & 0.01 & 0.24 & 0.00 & 0.11 & 0.15 & 0.10 \\
\midrule
\multirow{2}{*}{MASt3R-Fusion~\cite{zhou2026mast3rfusion}} & CD & 0.73 & \secondval{12.19} & 1.40 & 1.67 & 2.95 & 3.79 \\
 & $F_1$ & \bestval{44.34} & \bestval{20.31} & 19.08 & \bestval{27.24} & 15.78 & \secondval{25.35} \\
\midrule
\multirow{2}{*}{Ours w/ Sim(3)} & CD & \secondval{0.63} & \thirdval{12.20} & \secondval{0.77} & \thirdval{1.67} & \thirdval{1.02} & \thirdval{3.26} \\
 & $F_1$ & 36.46 & \thirdval{9.91} & \secondval{25.14} & \thirdval{20.00} & \thirdval{32.76} & \thirdval{24.85} \\
\multirow{2}{*}{Ours w/ ScaRF-SLAM~\cite{zhang2026scarfs}} & CD & \thirdval{0.64} & \bestval{11.83} & \thirdval{0.86} & \bestval{1.55} & \bestval{0.81} & \bestval{3.14} \\
 & $F_1$ & \thirdval{36.62} & 9.40 & \thirdval{23.81} & 19.33 & \secondval{32.86} & 24.40 \\
\multirow{2}{*}{\textbf{Ours}} & CD & \bestval{0.60} & 12.27 & \bestval{0.70} & \secondval{1.59} & \secondval{0.96} & \secondval{3.22} \\
 & $F_1$ & \secondval{43.34} & \secondval{11.67} & \bestval{30.16} & \secondval{23.48} & \bestval{36.78} & \bestval{29.09} \\
\bottomrule
\end{tabular*}
\end{table}
}

%% file: abstract.tex
\begin{abstract}
Upgrading feed-forward geometry to metric scale with visual-inertial odometry (VIO) raises the question of which correction should align dense feed-forward prediction with sparse metric VIO anchors.
Existing methods use either a similarity transformation, which cannot represent depth-dependent distortion, or a projective transformation, which alters pre-calibrated intrinsics.
We prove that the transformations compatible with pre-calibrated intrinsics form a ten-dimensional manifold.
Beyond Sim(3), this manifold contains three degrees of freedom (DoF) that locate the plane at infinity and leave residuals growing quadratically with depth range after any similarity alignment.
These DoF are invisible in the image and can be resolved only by metric depth along viewing rays or by multi-view constraints.
To estimate them, we propose \textbf{\method{}}, a probabilistic manifold optimization framework in which every anchor is weighted by its along-ray uncertainty and inlier probability within a single expectation-maximization (EM) procedure.
\method{} certifies the similarity synchronization and refines the plane at infinity only where it is observable, which guarantees an objective value no worse than that of the optimal similarity alignment.
Experiments on EuRoC, CID-SIMS, Oxford Spires, and KITTI-360 demonstrate that \method{} achieves the highest mean $F_1$ and outperforms Sim(3) alignment on all four datasets.
The project page is available at \url{https://h0xg.github.io/ProManO/}.
\end{abstract}


%% file: introduction.tex
\section{Introduction}
\label{sec:intro}
Dense metric reconstruction underpins robot navigation, simulation, and digital twins.
Recent 3D foundation models~\cite{wang2024dust3r,wang2025vggt,wang2025pi3,lin2026depthanything3,keetha2026mapanything} can predict dense, multi-view consistent geometry in a single forward pass, but do not provide reliable metric scale.
In contrast, visual-inertial odometry (VIO) provides metric camera motion and 3D structure, but only sparsely~\cite{qin2018vinsmono,geneva2020openvins,campos2021orbslam3}.
Combining the two offers a natural path to dense metric maps.
However, the key challenge lies in the metric upgrade itself: feed-forward geometry exhibits depth-dependent distortions that cannot be captured by a global similarity transformation.
Specifically, distant structure is unevenly compressed across the image, causing residual errors to increase with depth after similarity alignment to the VIO anchors (\cref{fig:teaser}), eventually leading to the rejection of distant anchors as outliers.

Existing approaches upgrade feed-forward geometry using correction models chosen by design rather than derived from geometric constraints.
Monocular depth pipelines estimate per-image scale and shift parameters~\cite{ranftl2020midas,wofk2023videpth,yin2021leres}.
Simultaneous localization and mapping (SLAM) systems built on feed-forward models align each submap using either a similarity transformation~\cite{murai2025mast3rslam,deng2025vggtlong,lee2026unisim}, which cannot capture depth-dependent distortions, or a projective transformation in SL(4)~\cite{maggio2025vggtslam,maggio2026vggtslam2}, whose additional degrees of freedom (DoF) may change pre-calibrated camera intrinsics.
Visual-inertial systems that fuse feed-forward geometry~\cite{zhou2026mast3rfusion,rhee2026foundationvio,salih2026vidar,lozano2026vi3} use metric measurements as alignment targets but retain these same correction
models.
However, none of these approaches derive the correction model from an analysis of the residual geometric ambiguity after camera calibration, nor do they explicitly account for the reliability of the available metric evidence.

\begin{figure}[t]
\centering
\includegraphics[width=\linewidth]{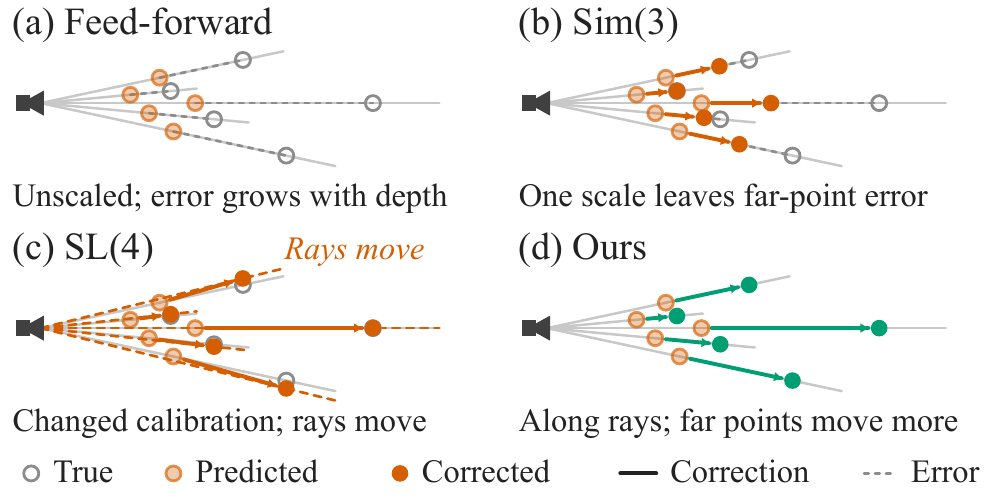}
\vspace{-4mm}
\caption{Correction models acting on predicted points.}
\label{fig:principle}
\vspace{-4mm}
\end{figure}

We model the residual error of feed-forward geometry as a transformation of the scene and prove that, with calibrated cameras, the intrinsics-consistent transformations form exactly a ten-dimensional manifold (\cref{thm:ambiguity}), namely a similarity and three modes that locate the plane at infinity.
These modes add the inverse-depth map of a plane to the prediction, are invisible in the image, bias any similarity alignment quadratically in depth (\cref{prop:bias}), and are resolved by metric depth along rays (\cref{fig:principle}).
This evidence is weakest where it is needed: a triangulated anchor is least precise along its viewing ray, wrong anchors cannot be exposed by the image, and constraints shared between submaps fix only their relative correction.
Building on this analysis, we present \textbf{\method{}} (\textbf{Pro}babilistic \textbf{Man}ifold \textbf{O}ptimization), which weights every anchor, overlap, and loop constraint by its along-ray uncertainty and inlier probability in one expectation-maximization (EM), certifies the similarity synchronization, and upgrades a submap from Sim(3) to the full correction only where an observability test shows that the data determine it.

In summary, our \textbf{contributions} are as follows:
\begin{itemize}
    \item
    We prove the intrinsics-consistent transformations are exactly Sim(3) and the position of the plane at infinity, derive its invisibility, depth-dependent bias, and degeneracies, and unify scale-shift, Sim(3), and SL(4) models.
    
    \item
    We propose \method{}, a probabilistic manifold optimization framework with along-ray anchor uncertainty, a single EM with monotone block updates across all stages, certified similarity synchronization, and an observability-gated refinement.
    
    \item 
    On four indoor and outdoor benchmarks, \method{} achieves the highest mean $F_1$ and improves $F_1$ over a certified similarity with the same anchors on every outdoor sequence, by 9.8 points on average on Oxford Spires.
\end{itemize}

%% file: related_work.tex
\section{Related Work}
\label{sec:related}

\paragraph{Feed-forward geometry and its correction.}
Pointmap regression from image pairs~\cite{wang2024dust3r,leroy2024mast3r} has grown into multi-view models that predict cameras and dense geometry~\cite{wang2025vggt,wang2025pi3,lin2026depthanything3,keetha2026mapanything}, streaming and long-context
variants~\cite{wang2025cut3r,chen2026lingbot,zhang2026loger,xie2026scal3r}, and metric backends~\cite{wang2026amb3r}.
Their predictions are consistent within a window but must be corrected across windows.
SLAM systems built on them align submaps with a similarity~\cite{murai2025mast3rslam,deng2025vggtlong,lee2026unisim,hu2026ec3rslam,jeon2026aimslam,f2slam}.
For uncalibrated cameras, VGGT-SLAM~\cite{maggio2025vggtslam} estimates a 15-DoF homography between consecutive submaps and optimizes on SL(4).
The estimate is degenerate on planar scenes and all 15 DoF drift between loop closures.
VGGT-SLAM~2.0~\cite{maggio2026vggtslam2} removes both by requiring the frame shared by two submaps to have one pose and one calibration, which leaves a calibration correction and a scale between submaps and no projective term.
Monocular depth is aligned per image by a scale and shift of inverse depth~\cite{ranftl2020midas,yin2021leres,wofk2023videpth}.
The similarity and scale-shift models are chosen by design, and the projective models follow from the ambiguity of uncalibrated cameras~\cite{hartley2003multiple}.
We derive the model for calibrated cameras: the intrinsics-consistent family between Sim(3) and SL(4).
It keeps the three projective modes of SL(4), which locate the plane at infinity, and excludes the five calibration modes that VGGT-SLAM~2.0 retains.
The inverse-depth shift is one of the three, while a shift in depth bends planes.

\paragraph{Metric anchoring.}
Visual-inertial methods constrain learned depth through depth codes~\cite{zuo2021codevio}, scale-shift fitting~\cite{wofk2023videpth}, affine inverse-depth models~\cite{merrill2024abvins}, or multi-view stereo~\cite{xin2023simplemapping,wang2025mvsvio}.
With feed-forward geometry, MASt3R-Fusion and FoundationVIO fuse it with inertial factor graphs~\cite{zhou2026mast3rfusion,rhee2026foundationvio}, ScaRF-SLAM optimizes frame and submap depth scales~\cite{zhang2026scarfs}, and VIDAR and LiDAR-VGGT anchor it with visual-inertial or LiDAR-inertial odometry~\cite{salih2026vidar,wang2025lidarvggt}.
These systems treat metric measurements as alignment targets.
We model each anchor's along-ray uncertainty and inlier probability, which our analysis identifies as what determines the plane at infinity.

\paragraph{Robust and certifiable estimation.}
Mixture models connect to reweighted least squares through outlier processes and graduated non-convexity~\cite{black1996unification,yang2020gnc}.
Synchronization problems are solved to certified optimality by semidefinite relaxation~\cite{zhao2026advances,liao2025convex,liao2024globalpointer} with Burer--Monteiro factorization~\cite{burer2003nonlinear,boumal2016nonconvex,rosen2019sesync,dellaert2020shonan}, including anisotropic rotation averaging~\cite{olsson2025anisotropic}, similarity synchronization~\cite{yu2024simsync}, and bundle adjustment with learned depth~\cite{han2025rome}.
We extend certifiable similarity synchronization with metric anchors, anisotropic covariances, and a chordal prior, embed it in an EM whose weights are those of an outlier process, and add an observability test that decides where the plane at infinity may be estimated.

%% file: residual_ambiguity.tex
\section{Intrinsics-Consistent Ambiguity}
\label{sec:ambiguity}

We model the residual error of a feed-forward prediction as an unknown projective transformation of the true scene and ask which transformations remain possible once the cameras are calibrated.

\PAR{Setting.}
Camera $f$ has intrinsics $\mathbf K_f$, center $\mathbf c_f$, and pose $\mathbf T_f$, so $P_f=\mathbf K_f[\mathbf I_3\,|\,\mathbf 0]\mathbf T_f^{-1}$.
A pixel has normalized coordinates $\mathbf b=\mathbf K_f^{-1}\widetilde{\mathbf u}=(x,y,1)^\top$ and predicted inverse depth $\rho=1/z$. Uncalibrated geometry is determined only up to $H\in PGL(4)$, with 15 DoF~\cite{hartley2003multiple,maggio2025vggtslam}.
We seek the transformations compatible with the known calibration:
\begin{equation}
  \mathcal M_f=\big\{H:\ P_fH^{-1}\simeq\mathbf K_f[\mathbf R'\,|\,\mathbf t'],\
  \mathbf R'\in SO(3)\big\}.
  \label{eq:camera_consistent}
\end{equation}
Let $\mathrm{Stab}(P_f)=\{H:P_fH^{-1}\simeq P_f\}$, $\mathcal H_f$ be the homotheties about $\mathbf c_f$, and $\mathcal Q_f$ be the elations centered at $\mathbf c_f$:
\begin{equation}
  \mathcal Q_f=\big\{\mathbf T_fE_{\mathbf q}\mathbf T_f^{-1}\big\}_{\mathbf q\in\mathbb R^3},
  \quad
  E_{\mathbf q}=\left[\begin{smallmatrix}\mathbf I_3&\mathbf 0\\ \mathbf q^\top&1\end{smallmatrix}\right].
  \label{eq:elation}
\end{equation}

\begin{theorem}[Intrinsics-consistent ambiguity]
\label{thm:ambiguity}
(i) $\mathrm{Stab}(P_f)=\mathcal H_f\ltimes\mathcal Q_f$ is the four-dimensional group fixing $\mathbf c_f$ and every ray through it, with $\mathcal Q_f\cong(\mathbb R^3,+)$.
(ii) An elation shifts the inverse depth of camera $f$ additively, $\rho'=\rho+\mathbf q^\top\mathbf b$.
(iii) $\mathcal M_f=\mathrm{Sim}(3)\cdot\mathcal Q_f$ with unique factorization.
It is a ten-dimensional immersed submanifold of $PGL(4)$.
\end{theorem}

\begin{lemma}[Sim(3) misses exactly three modes]
\label{lem:gap}
$\mathrm{Sim}(3)\cap\mathcal Q_f=\{\mathbf I\}$ and
$\mathrm{Sim}(3)\cap\mathrm{Stab}(P_f)=\mathcal H_f$.
\end{lemma}

\PAR{The plane at infinity.}
Of the 15 DoF of $PGL(4)$, 7 form a similarity, 5 change the intrinsics, and 3 move the plane at infinity $\pi_\infty$~\cite{hartley2003multiple}.
Calibration removes the 5, so the ambiguity beyond Sim(3) is exactly the location of $\pi_\infty$: $E_{\mathbf q}$ maps it to the plane $\mathbf q^\top\mathbf X=1$ and compresses distant structure toward depth $1/(\mathbf q^\top\mathbf b)$.
Here $q_3$ is the familiar inverse-depth shift, and $q_1,q_2$ tilt it across the image (\cref{fig:principle}).
The offset must act on inverse depth: a depth offset $z'=z+\beta$ bends planes and lies outside $\mathcal M_f$.

\begin{proposition}[Invisibility]
\label{prop:invisible}
Elations centered at $\mathbf c_f$ leave every image of camera $f$ unchanged.
Only metric depth along its rays or other viewpoints can resolve them.
\end{proposition}

\begin{proposition}[Similarity alignment is biased]
\label{prop:bias}
For three collinear points spaced $L/2$ apart along a unit direction $\mathbf v$ and distorted by $E_{\mathbf q}$, every similarity alignment leaves, to first order, a residual of at least $|\mathbf q^\top\mathbf v|L^2/8$.
\end{proposition}
\noindent More anchors cannot remove this bias, so Sim(3) alone rejects distant anchors.
\cref{sec:submap_alignment} marginalizes $\mathbf q$ to prevent this.

\PAR{Submap model.}
The modes beyond Sim(3) are tied to one optical center: no transformation other than the identity fixes every ray of two cameras with distinct centers.
We therefore parameterize a submap by one similarity $\mathbf S_c$ shared by its frames and coefficients $\mathbf q_c$ applied in each frame along its own rays.
This assumes that the feed-forward bias is camera-centric and stationary within one forward pass.
Keyframes are selected by rotational parallax, which limits how far the viewing direction changes within a submap.
Overlaps tie neighboring submaps.
$\mathbf q_c$ is introduced only after the similarity is certified, and only where it is observable (\cref{sec:global_alignment}).

%% file: method.tex
\input{method/00_overview}

\LayoutIndoorAte
\input{method/02_visual_inertial_anchors}
\LayoutQualitative
\LayoutOutdoorAte
\input{method/03_local_alignment}
\LayoutCloseups
\LayoutEurocRecon
\LayoutCidRecon
\input{method/04_global_alignment}

%% file: method/00_overview.tex
\section{Probabilistic Manifold Optimization}
\label{sec:method}

Given images and inertial measurement unit (IMU) data, a VIO frontend provides metric poses, intrinsics, and sparse landmarks, which stay fixed.
Keyframes form overlapping submaps, for which a feed-forward model predicts dense point maps from the images alone.
For each submap, we estimate the intrinsics-consistent correction of \cref{sec:ambiguity} in three modules (\cref{fig:pipeline}): the visual-inertial frontend turns landmarks into metric anchors with along-ray uncertainty and inlier probability (\cref{sec:vi_anchors}).
Submap-level probabilistic optimization aligns each submap by a similarity with $\mathbf q$ marginalized (\cref{sec:submap_alignment}).
Global intrinsics-consistent optimization certifies the similarity synchronization, gates $\mathbf q$ by observability, and refines all states by EM (\cref{sec:global_alignment}).

%% file: method/02_visual_inertial_anchors.tex
\subsection{Visual-Inertial Anchors with Along-Ray Uncertainty}
\label{sec:vi_anchors}

By ~\cref{sec:ambiguity}, the modes beyond Sim(3) require metric depth along rays, so the frontend must deliver not only metric motion but also anchors with inlier probabilities and along-ray uncertainties.
We build it on VINS-Mono~\cite{qin2018vinsmono}, with its IMU preintegration and sliding window, in three steps (\cref{fig:pipeline}, upper left):
(i) tracks receive reliability priors, (ii) the sliding-window estimator is weighted by these priors, and (iii) the converged geometry is turned into anchors by fixed-pose re-triangulation.

\PAR{KLT tracking and reliability priors.}
Each Kanade-Lucas-Tomasi (KLT) observation of landmark $l$ in frame $j$ receives an image-based prior $p_{lj}\in[0,1]$ from its forward-backward residual and patch correlation, and a posterior that adds geometric consistency through a Gaussian mixture on its reprojection residual $\mathbf e_{lj}$ under the current geometry, with $\mathcal N_\bullet(\mathbf e)=\mathcal N(\mathbf e;\mathbf 0,\boldsymbol\Sigma_\bullet)$:
\begin{equation}
  \gamma^{\mathrm{trk}}_{lj}=
  \frac{p_{lj}\,\mathcal N_I(\mathbf e_{lj})}
       {p_{lj}\,\mathcal N_I(\mathbf e_{lj})+(1-p_{lj})\,\mathcal N_O(\mathbf e_{lj})}.
  \label{eq:trk_posterior}
\end{equation}

\PAR{Posterior-weighted sliding-window optimization.}
The additional tracks enter the VINS-Mono objective as reprojection factors weighted by $\gamma^{\mathrm{trk}}_{lj}$, next to its marginalization, IMU, and original feature terms.
Reliability and geometry are estimated alternately: the posteriors in \cref{eq:trk_posterior} weight the window, and the optimized window updates the residuals $\mathbf e_{lj}$ and hence the posteriors.
The converged window provides the metric poses $\mathbf T_f$ and the landmark tracks used below.

\PAR{Metric anchor construction.}
With the poses fixed, each landmark is re-triangulated with posterior weights (fixed-pose re-triangulation):
\begin{equation}
  \mathbf X_h=\arg\min_{\mathbf X}\sum\nolimits_{j\in\mathcal O_h}
  \gamma^{X}_{hj}\,
  \psi\!\big(\|\pi(\mathbf T_{cw,j}\mathbf X)-\mathbf u_{hj}\|^2\big),
  \label{eq:anchor_refine}
\end{equation}
and its covariance follows from the information matrix:
\begin{equation}
  \boldsymbol\Sigma^{s}_h=(\mathcal I_h+\epsilon\mathbf I_3)^{-1}+\sigma_p^2\mathbf I_3,
  \label{eq:anchor_cov}
\end{equation}
with the information matrix $\mathcal I_h=\sum_{j\in\mathcal V_h}\gamma^{X}_{hj}\mathbf J_{hj}^{\top}\mathbf W_{hj}\mathbf J_{hj}$ where $\mathbf W_{hj}=(p_{hj}/\sigma_{\mathrm{px}}^2)\mathbf I_2$, verified observations $\mathcal V_h$, a regularizer $\epsilon$, and a floor $\sigma_p^2$ for pose uncertainty.
Because the plane at infinity acts only along rays, for a camera $f$ observing $\mathbf X_h$ at $\mathbf b_{h,f}$ with depth $z_{h,f}$ and unit ray $\mathbf n_{h,f}$, we split $\boldsymbol\Sigma^s_h$ into an along-ray inverse-depth variance and a transverse covariance:
\begin{equation}
  \sigma^2_{\rho,h}(f)=\frac{\mathbf n_{h,f}^{\top}\boldsymbol\Sigma^{s}_h\mathbf n_{h,f}}
    {(\|\mathbf b_{h,f}\|z_{h,f}^{2})^{2}},\;\;
  \boldsymbol\Sigma^{\perp}_h(f)=\mathbf P^{\perp}_{h,f}\boldsymbol\Sigma^{s}_h\mathbf P^{\perp}_{h,f},
  \label{eq:axial_transverse}
\end{equation}
with $\mathbf P^{\perp}_{h,f}=\mathbf I_3-\mathbf n_{h,f}\mathbf n_{h,f}^{\top}$.
Inverse depth keeps this uncertainty near-Gaussian even at low parallax~\cite{civera2008inverse}.
Each anchor is passed to the backend as $(\mathbf X_h,\pi_h,\boldsymbol\Sigma^s_h)$, where the inlier prior $\pi_h=|\mathcal O_h|^{-1}\sum_{j\in\mathcal O_h}\gamma^{X}_{hj}$ averages the posteriors of its observations.
These components describe the directional uncertainty of the anchor. The backend uses its full covariance $\boldsymbol\Sigma^s_h$.
Poorly triangulated anchors thus inform the similarity but barely the plane at infinity.

%% file: method/03_local_alignment.tex
\subsection{Submap-Level Probabilistic Optimization}
\label{sec:submap_alignment}
\label{sec:local_alignment}%

Each submap is processed in four steps (\cref{fig:pipeline}, upper right): (i) the feed-forward model predicts its geometry, (ii) the prediction is calibrated and aligned to the VIO frame, (iii) every point and anchor receives an uncertainty, and (iv) the similarity is estimated by EM with the plane at infinity marginalized.

\PAR{Feed-forward reconstruction.}
Each submap $\mathcal K_c$ contains new keyframes, the keyframes shared with submap $\mathcal K_{c-1}$, and verified loop keyframes.
From the images alone, a feed-forward model predicts for every frame a point map $\hat{\mathbf P}_{c,f}$ with confidence $Q_{c,f}$ and a camera pose $\hat{\mathbf T}_{c,f}$ in a submap frame.
For a pixel $j=(f,\mathbf u)$, let $\mathbf X_{c,j}=\hat{\mathbf T}_{c,f}^{-1}\hat{\mathbf P}_{c,f}(\mathbf u)$ be its corresponding 3D point in the coordinates of camera $f$, with depth $z_{c,j}=\mathbf e_3^\top\mathbf X_{c,j}$, normalized coordinates $\mathbf b_j=\mathbf X_{c,j}/z_{c,j}$, and unit ray
$\mathbf n_{c,j}=\hat{\mathbf R}_{c,f}\mathbf b_j/\|\mathbf b_j\|$.
The intrinsics-consistent correction of \cref{sec:ambiguity} acts on these camera-frame points and needs no intrinsics.
Deviations of the predicted cameras from the calibration are visible in the images and enter the residual noise.

\PAR{Metric initialization.}
Anchors observed in $\mathcal K_c$ give correspondences $h\in\mathcal A_c$ at pixels $j(h)=(f_h,\mathbf u_h)$.
They first fix the per-ray homothety in closed form, $\alpha_c=\operatorname{median}_{h\in\mathcal A_c} z^{s}_{h}/z_{c,j(h)}$,
with $z^s_h$ the depth of $\mathbf X_h$ in VIO camera $f_h$, which calibrates the prediction to:
\begin{equation}
  z^{0}_{c,j}=\alpha_c z_{c,j},\qquad
  \mathbf P^{0}_{c,j}=\hat{\mathbf R}_{c,f}z^{0}_{c,j}\mathbf b_j+\hat{\mathbf t}_{c,f},
  \label{eq:ffd_backprojection}
\end{equation}
with inverse depth $\rho^0_{c,j}=1/z^0_{c,j}$.
The similarity $\mathbf S_c$ then maps the submap frame to the VIO world frame.
$\mathbf S_c$ and $\mathbf q_c$ are the only free variables of a submap.

\PAR{Uncertainty modeling.}
Each anchor $h$ and its dense point $\mathbf P^0_{c,j(h)}$ lie on one ray.
The dense point has a surface-oriented covariance $\mathbf C^d_{c,j}$ whose normal variance grows with depth, low confidence, and grazing incidence.
The anchor is precise across its ray but not along it (\cref{sec:vi_anchors}).
In the local stage, $\mathbf q_c$ is weakly constrained, so we marginalize it with the prior $\mathbf q_c\sim\mathcal N(\mathbf 0,\mathbf V_0)$, $\mathbf V_0=v_{\max}\mathbf I_3$.
Since $\mathbf q_c$ moves points only along their rays, this inflates the covariance to:
\begin{equation}
  \tilde{\mathbf C}^{d}_{c,j}=\mathbf C^{d}_{c,j}
  +\|\mathbf b_j\|^{2}(z^{0}_{c,j})^{4}
   (\mathbf b_j^{\top}\mathbf V_0\mathbf b_j)\,
   \mathbf n_{c,j}\mathbf n_{c,j}^{\top},
  \label{eq:marginal_cov}
\end{equation}
which grows with depth, so distant anchors are not rejected for the bias of \cref{prop:bias}.
Once $\mathbf q_c$ is estimated (\cref{sec:global_alignment}), the inflation is dropped.

\PAR{Local probabilistic alignment.}
Each anchor correspondence contributes a residual.
With similarity $\mathbf S_c=(s_c,\mathbf R_c,\mathbf t_c)$, fixed pivot $\mathbf o_c$, %
and noise scale $\nu_c$:
\begin{equation}
  \begin{aligned}
  \mathbf d_{ch}&=\mathbf o_c+\mathbf t_c
   +s_c\mathbf R_c\big(\mathbf P^{0}_{c,j(h)}-\mathbf o_c\big)-\mathbf X_h,\\
  \boldsymbol\Sigma_{ch}&=\nu_c^{2}\big(\boldsymbol\Sigma^{s}_h
   +s_c^{2}\mathbf R_c\tilde{\mathbf C}^{d}_{c,j(h)}\mathbf R_c^{\top}\big),
  \end{aligned}
  \label{eq:metric_discrepancy}
\end{equation}
so the residual is weighted by the sum of the anchor and dense covariances: tolerant along the ray, where neither is precise, and strict across it. Wrong correspondences are handled by the mixture:
\begin{equation}
  \mathbf d_{ch}\sim\pi_h\,\mathcal N(\mathbf 0,\boldsymbol\Sigma_{ch})
  +(1-\pi_h)\,\mathcal N(\mathbf 0,\chi^{2}\boldsymbol\Sigma_{ch}),
  \label{eq:metric_mixture}
\end{equation}
with the anchor prior $\pi_h$ of \cref{sec:vi_anchors} and $\chi>1$.
Each submap is then aligned independently by EM~\cite{dempster1977em}.
The E-step weights the correspondences:
\begin{equation}
  \gamma^{(k)}_{ch}=\operatorname{sigmoid}\Big(
  \operatorname{logit}(\pi_h)+3\log\chi
  -\tfrac{1-\chi^{-2}}{2}M^{(k)}_{ch}\Big),
  \label{eq:local_estep}
\end{equation}
with $M^{(k)}_{ch}=\|\mathbf d^{(k)}_{ch}\|^{2}_{(\boldsymbol\Sigma^{(k)}_{ch})^{-1}}$.
The M-step updates the submap similarity, with weights $w^{(k)}_{ch}=\gamma^{(k)}_{ch}+(1-\gamma^{(k)}_{ch})/\chi^{2}$ and a prior on the similarity coordinates $\boldsymbol\xi_c$:
\begin{equation}
  \begin{aligned}
  \mathbf S_c^{(k+1)}=\arg\min_{\mathbf S_c}
  &\sum_{h\in\mathcal A_c}w^{(k)}_{ch}
   \|\mathbf d_{ch}\|^{2}_{(\boldsymbol\Sigma^{(k)}_{ch})^{-1}}\\
  &+\|\boldsymbol\xi_c-\boldsymbol\xi_{c,0}\|^{2}_{\boldsymbol\Lambda_{c,0}},
  \end{aligned}
  \label{eq:local_mstep}
\end{equation}
solved by Levenberg--Marquardt~\cite{levenberg1944method,marquardt1963algorithm}.
The noise scale $\nu_c$ is updated in closed form from the weighted Mahalanobis residuals.
With the proxies $\bar{\boldsymbol\Sigma}_{ch}$ fixed within each outer round, these updates are monotone.
The result $\mathbf S_c^{\mathrm{loc}}$ initializes \cref{sec:global_alignment}.

%% file: method/04_global_alignment.tex
\subsection{Global Optimization on the Intrinsics-Consistent Manifold}
\label{sec:global_alignment}
\label{sec:global}%

The global stage couples all submaps (\cref{fig:pipeline}, bottom): (i) builds a factor graph over their ten-dimensional states, (ii) solves the similarity synchronization to certified optimality, (iii) estimates the plane at infinity where it is observable, and (iv) applies the corrections to every pixel.

\paragraph{Global factor graph.}
Each submap carries $(\mathbf S_c,\mathbf q_c)\in\mathrm{Sim}(3)\times\mathbb R^3$, with $\mathbf q_c$ in $\mathrm m^{-1}$ shared by its frames.
The corrected inverse depth of pixel $j=(f,\mathbf u)$ is $\rho'_{c,j}=\rho^{0}_{c,j}+\mathbf q_c^{\top}\mathbf b_j$, and its metric point:
\begin{equation}
  \mathbf F_c(j)=\mathbf o_c+\mathbf t_c
  +s_c\mathbf R_c\big(\hat{\mathbf R}_{c,f}\mathbf b_j/\rho'_{c,j}
   +\hat{\mathbf t}_{c,f}-\mathbf o_c\big),
  \label{eq:plane_correction}
\end{equation}
and depth stays positive by restricting $\mathbf q_c$ to the polyhedron $\mathcal D_c$ on which $\rho'_{c,j}\ge\eta\rho^0_{\min,c}$ at every pixel, four linear constraints at the image corners since $\mathbf b^\top\mathbf q$ is affine in the image.
Three kinds of factors connect the states. 
Anchor factors $e=h$ reuse the local correspondences, while overlap and loop factors $e=(j_a,j_b)$ pair the predictions of the same pixel of a keyframe shared by submaps $a$ and $b$.
With $\mathbf C_c(j)=s_c^{2}\mathbf R_c\mathbf C^{d}_{c,j}\mathbf R_c^{\top}$,
with every term evaluated at the factor's pixel ($j(h)$ for an anchor, $j_a,j_b$ for a pair), the residuals and covariances are:
\begin{equation}
  \begin{aligned}
  \mathbf d_e&=\mathbf F_c-\mathbf X_h,&
   \boldsymbol\Sigma_e&=\nu_{\mathrm A}^{2}\big(\boldsymbol\Sigma^{s}_h+\mathbf C_c\big)
   &&\text{(anchor)},\\
  \mathbf d_e&=\mathbf F_a-\mathbf F_b,&
   \boldsymbol\Sigma_e&=\nu_{g}^{2}\big(\mathbf C_a+\mathbf C_b\big)
   &&\text{(pair)},
  \end{aligned}
  \label{eq:global_factors}
\end{equation}
with $g\in\{\mathrm O,\mathrm L\}$, and $\tilde{\mathbf C}^d$ replacing $\mathbf C^d$ for submaps whose $\mathbf q_c$ is not estimated.
Each factor follows the mixture in~\cref{eq:metric_mixture}, with prior $\pi_h$ for anchors
and $\pi_{\mathrm O},\pi_{\mathrm L}$ for pairs,
and mass weights $\beta_e$ cap redundant correspondences. The Jacobian:
\begin{equation}
  \partial\mathbf F_c(j)/\partial\mathbf q_c
  =-\big(z'_{c,j}\big)^{2}s_c\mathbf R_c\hat{\mathbf R}_{c,f}\,
   \mathbf b_j\mathbf b_j^{\top}
  \label{eq:q_jacobian}
\end{equation}
has rank one along the ray, so $\mathbf q_c$ is observed only through along-ray residuals, and pairwise factors on off-plane structure can resolve the ambiguity between scale and ray correction left by coplanar anchors.
With $w_e=\gamma_e+(1-\gamma_e)/\chi^{2}$ and $\mathbf G_c=s_c\mathbf R_c$, the graph defines the objective $\Phi(\mathbf S,\mathbf q)$:
\begin{equation}
  \Phi=\tfrac12\Big(\sum\nolimits_{e}\beta_e w_e
   \|\mathbf d_e\|^{2}_{\boldsymbol\Sigma_e^{-1}}
   +\sum\nolimits_{\mathcal U}\|\mathbf q_c\|^{2}_{\mathbf V^{-1}}
   +\Psi\Big),
  \label{eq:global_objective}
\end{equation}
where $\mathcal U$ is the set of submaps whose $\mathbf q_c$ is estimated ($\mathbf q_c=\mathbf 0$ otherwise), $\mathbf V=\mathrm{diag}(v_1,v_2,v_3)$ holds adaptive prior variances, and the chordal prior $\Psi=\sum_c\big(\|\mathbf G_c-\mathbf G^{0}_c\|_F^{2}/\sigma_{\mathrm{lin}}^{2}+\|\mathbf t_c-\mathbf t^{0}_c\|^{2}/\sigma_{\mathrm{tr}}^{2}\big)$, centered at $\mathbf S^{\mathrm{loc}}_c$, keeps $\Phi$ quadratic in $(\mathbf G_c,\mathbf t_c)$ for fixed $\mathbf q$.

\LayoutSequenceMaps
\PAR{Certified similarity alignment.}
With $\mathcal U=\emptyset$, we alternate the E-step, the similarity update, and the noise update.
For fixed $\mathbf q$ and covariance proxies, the similarity update is a quadratically constrained quadratic program (QCQP) in $\{\mathbf G_c\}$ with $\mathbf G_c^\top\mathbf G_c\propto\mathbf I_3$.
We solve its semidefinite relaxation by Burer--Monteiro factorization with a Riemannian staircase and verify global optimality with a dual certificate~\cite{rosen2019sesync,yu2024simsync}.

\PAR{Intrinsics-consistent correction.}
To model depth distortions beyond similarity alignment, we update the plane-at-infinity parameter $\mathbf q_c$ while preserving camera intrinsics.
From the data factors alone, with unmarginalized covariances and no priors, the reduced information:
\begin{equation}
  \mathbf H_{\mathbf q|\mathbf S}=\mathbf H_{\mathbf q\mathbf q}
  -\mathbf H_{\mathbf q\mathbf S}\mathbf H_{\mathbf S\mathbf S}^{-1}
   \mathbf H_{\mathbf S\mathbf q}
  \label{eq:reduced_information}
\end{equation}
measures what the data reveal about $\mathbf q$ once the similarity has absorbed what it can.
Since overlaps and loops constrain only differences of the $\mathbf q_c$, we gate on the marginal information over the factor graph, $\mathcal U=\{c:\lambda_{\min}([(\mathbf H_{\mathbf q|\mathbf S})^{-1}]_{cc}^{-1})\ge\tau_{\mathrm{obs}}\}$, $\tau_{\mathrm{obs}}=\varsigma/v_{\max}$ with $\varsigma\ge1$, so that the data dominate the prior in every released direction.
For the remaining submaps, $\mathbf q_c$ stays at zero while $\mathbf S_c$ is still updated jointly with the other similarities.
For $c\in\mathcal U$, with the proxies frozen, we cycle the E-step, the certified similarity update (first at $\mathbf q=\mathbf 0$), a damped Gauss--Newton step on $\mathbf q$ (a convex quadratic program (QP) over $\{\mathcal D_c\}$), and closed-form updates of $\nu_g$ and $v_r$.

\begin{proposition}[Safe refinement]
\label{prop:safe}
With $\mathcal U$ and the proxies fixed, no refinement update increases the negative log-posterior, so the final estimate is no worse than $\bar{\mathbf S}=\arg\min_{\mathbf S}\Phi(\mathbf S,\mathbf 0)$, the optimal similarity-only alignment at the start of the refinement.
\end{proposition}

\noindent %
The gate keeps $\mathbf q$ from fitting noise where it is unobservable.

\PAR{Apply corrections and fuse.}
Finally, \cref{eq:plane_correction} is applied to every pixel of every submap with the estimated $(\mathbf S_c,\mathbf q_c)$, which places all corrected points in the VIO world frame, and the submaps are fused into the final metric map.

%% file: experiments.tex
\section{Experiments}
\label{sec:experiments}

\LayoutOxfordRecon
\LayoutKittiRecon
\LayoutOutdoorSequenceMaps
\subsection{Experimental Setup}
\PAR{Datasets.}
We evaluate localization and dense reconstruction on 28 sequences: six from EuRoC~\cite{burri2016euroc}, eight from CID-SIMS~\cite{zhang2024cidsims}, nine from Oxford Spires~\cite{tao2026oxfordspires}, and five from KITTI-360~\cite{liao2022kitti360}, covering aerial motion, indoor robot navigation, large outdoor environments, and urban driving.

\PAR{Baselines.}
We separate vision-only (VO) and VIO methods by their sensor inputs.
The VO group comprises LingBot-Map~\cite{chen2026lingbot}, LoGeR~\cite{zhang2026loger}, AMB3R~\cite{wang2026amb3r}, MASt3R-SLAM~\cite{murai2025mast3rslam}, and streaming Depth Anything~3 (DA3)~\cite{lin2026depthanything3}.
The VIO group includes ORB-SLAM3~\cite{campos2021orbslam3}, OpenVINS~\cite{geneva2020openvins}, VINS-Mono~\cite{qin2018vinsmono}, VINS-Fusion~\cite{qin2019vinsfusion}, Kimera-VIO~\cite{rosinol2020kimera}, OKVIS2~\cite{leutenegger2022okvis2}, DM-VIO~\cite{stumberg2022dmvio}, and MASt3R-Fusion~\cite{zhou2026mast3rfusion}.
For LoGeR, LingBot-Map, AMB3R, and streaming DA3, we use the keyframes selected by our frontend as input. Under this shared-input protocol, numbers are not directly comparable to the original papers.
Dense reconstruction is compared against VO methods and MASt3R-Fusion.
We additionally compare our reconstruction backend with Sim(3) alignment and a ScaRF-SLAM-based mapping variant~\cite{zhang2026scarfs}, using the same frontend trajectory.
Unless otherwise specified, the \bestval{best}, \secondval{second-best}, and \thirdval{third-best} results are highlighted in bold, underlined, and italicized text, respectively.

\PAR{Metrics.}
Localization is measured by the root mean square error of absolute trajectory error (ATE RMSE) in meters.
We use Sim(3) alignment for VO methods to account for its scale ambiguity and SE(3) alignment for VIO methods to preserve metric-scale errors.
For reconstruction, maps are aligned to the ground truth with the same convention as the trajectories, \ie, Sim(3) for VO methods and SE(3) for VIO methods, and are then evaluated with identical downsampling and scoring settings for all methods within each benchmark.
We use a ScaRF-SLAM-based nearest-neighbor evaluation protocol~\cite{zhang2026scarfs}, using thresholds of $5$\,cm for indoors and $20$\,cm for outdoors.
Chamfer distance (CD, in meters) averages the mean prediction-to-reference distance (accuracy, Acc.)\ and the mean reference-to-prediction distance.
$F_1$ (in percent) is the harmonic mean of the fractions of points within the threshold in the two directions.
A run is marked $\times$ if it returns no trajectory, covers less than $80\%$ of the sequence, or diverges (ATE above that of a stationary estimate).
An average is marked $\times$ if any of its sequences is tracking failure.

\PAR{Implementation details.}
Our dense predictor is DA3NESTED-GIANT-LARGE-1.1.
Submaps contain up to 32 keyframes with four shared keyframes.
The backend corrects their geometry while keeping the frontend poses fixed.
The feed-forward pass takes $5.1$\,s per submap on one RTX~4070 GPU, and the local and global optimization, which run on an Intel Core i7-12700KF CPU, add $0.9$\,s per submap.

\subsection{Localization Accuracy}
\cref{tab:ate_indoor,tab:ate_outdoor} evaluate the metric trajectories used to anchor reconstruction.
Our average ATE is $0.084$\,m on EuRoC, $0.268$\,m on CID-SIMS, $0.39$\,m on Oxford Spires, and $4.58$\,m on KITTI-360.
These are the lowest ATE results on CID-SIMS, Oxford Spires, and KITTI-360.
\Cref{fig:indoor_sequence_maps,fig:outdoor_sequence_maps} show the estimated camera paths in the corresponding full-sequence point clouds.

\subsection{Dense Reconstruction}
\cref{tab:recon_euroc,tab:recon_cid,tab:recon_oxford,tab:recon_kitti} assess geometric accuracy and completeness.
Our method achieves the highest reported mean $F_1$ on all four datasets: $53.99\%$ on EuRoC, $45.24\%$ on CID-SIMS, $60.74\%$ on Oxford Spires, and $29.09\%$ on KITTI-360. It also attains the lowest mean CD on EuRoC, CID-SIMS, and Oxford Spires ($0.065$, $0.179$, and $0.37$\,m).

\PAR{Reconstruction backend comparison.}
With the frontend trajectory fixed, our method improves mean $F_1$ over the Ours w/ Sim(3) variant by $4.1$ to $9.8$ points and over the Ours w/ ScaRF-SLAM variant by $4.7$ to $10.4$ points across the four datasets.
\cref{sec:ablation} ablates the components.

\input{tables/tab_model_alignment}
\PAR{Feed-forward model comparison.}
\cref{tab:model_alignment} replaces the feed-forward model. 
The correction improves mean $F_1$ over Sim(3) alignment (Sim(3)) for all four models on all four datasets, by $2.2$ to $9.8$ points, so the gain does not depend on a particular backbone.

\PAR{Qualitative results.}
\cref{fig:qualitative_reconstruction} compares EuRoC V1\_03 and CID-SIMS apartment1\_1, while \cref{fig:region_closeups} provides indoor and outdoor close-ups for inspecting surface continuity and missing structure, alongside the quantitative metrics.

\PAR{Indoor sequence maps.}
The top-down views in \cref{fig:indoor_sequence_maps} show repeated passes through the EuRoC rooms. On CID-SIMS, the 14-13-12 route forms a compact loop, while the Floor~14 route links several parallel corridors. These full maps put the room-scale errors in \cref{fig:qualitative_reconstruction} in the context of the camera motion.

\PAR{Outdoor sequence maps.}
\Cref{fig:outdoor_sequence_maps} shows the two connected courtyards at Keble College and the street network covered by each KITTI-360 drive. Drive~0000 follows a long main road with several side streets, drive~0005 includes a long curved segment, and drive~0009 visits multiple adjoining blocks. The full maps show how these parts of each route fit together, while the matched views in \cref{fig:region_closeups} compare local surface errors.

\begin{table}[t]
\centering
\captionsetup{font=small}
\caption{Ablation study. Average CD [m] $\downarrow$ and $F_1$ [\%] $\uparrow$ at $5$\,cm (CID-SIMS~\cite{zhang2024cidsims}) and $20$\,cm (Oxford Spires~\cite{tao2026oxfordspires}) are reported.}
\vspace{-2mm}
\label{tab:ablation}
\footnotesize
\setlength{\tabcolsep}{3pt}
\resizebox{\linewidth}{!}{%
\begin{tabular*}{1.25\linewidth}{@{}l@{\extracolsep{\fill}}lrrrr@{}}
\toprule
 & & \multicolumn{2}{c}{CID-SIMS} & \multicolumn{2}{c}{Oxford Spires} \\
\cmidrule(lr){3-4} \cmidrule(lr){5-6}
Module & Variant & CD & $F_1$ & CD & $F_1$ \\
\midrule
VI anchors & Isotropic anchor covariance & 0.235 & 36.56 & 0.85 & 44.35 \\
\midrule
Submap opt. & w/o marginalizing $\mathbf q_c$ & 0.186 & \underline{43.68} & 0.65 &\underline{56.32} \\
\midrule
\multirow{4}{*}{Global opt.} & w/o $\mathbf q_c$ ($\mathbf q_c=\mathbf 0$) & 0.195 & 42.66 & \underline{0.45} & 53.22 \\
 & Only $q_{c,3}$ & \underline{0.185} & 43.59 & 0.46 & 53.68 \\
 & Per-frame $\mathbf q$ & 0.223 & 38.93 & 0.69 & 41.56 \\
 & w/o gate & 0.186 & 42.22 & 0.63 & 46.36 \\
\midrule
\multicolumn{2}{@{}l}{Ours w/ Sim(3) alignment} & 0.199 & 40.49 & 0.51 & 50.97 \\
\multicolumn{2}{@{}l}{\textbf{Ours}} & \textbf{0.179} & \textbf{45.24} & \textbf{0.37} & \textbf{60.74} \\
\bottomrule
\end{tabular*}}
\vspace{-5mm}
\end{table}

\subsection{Ablation Study}
\label{sec:ablation}
\begin{figure}[t]
\centering
\includegraphics[width=\linewidth]{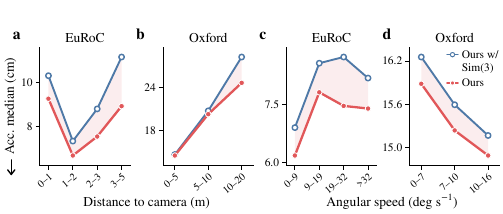}
\vspace{-4mm}
\caption{Median accuracy [cm] $\downarrow$ of \method{} and Ours w/ Sim(3) variant, binned by distance to the camera  and by camera angular speed .}
\vspace{-4mm}

\label{fig:acc_depth_speed}
\end{figure}

\cref{tab:ablation} changes one component at a time. Ours w/ Sim(3) alignment, the similarity-only variant of \cref{tab:recon_euroc,tab:recon_cid,tab:recon_oxford,tab:recon_kitti}, stops after the similarity synchronization, whereas $\mathbf q_c=\mathbf 0$ still runs the EM refinement of the global optimization with $\mathbf q_c$ fixed at zero.
For the VI anchors, an isotropic anchor covariance of equal trace in place of \cref{eq:anchor_cov} lets poorly triangulated depth constrain $\mathbf q_c$ and loses $8.7$ $F_1$ points on CID-SIMS and $16.4$ on Oxford Spires.
In the submap optimization, dropping the along-ray inflation of \cref{eq:marginal_cov} fits the similarity as if the prediction had no far-field bias, and $F_1$ drops by $1.6$ and $4.4$ points.
In the global optimization, the full $\mathbf q_c$ gains $2.6$ and $7.5$ points over $\mathbf q_c=\mathbf 0$, but the inverse-depth offset $q_{c,3}$ alone gains less than one, so most of the gain comes from the tilt modes $q_{c,1}$ and $q_{c,2}$.
One $\mathbf q$ per keyframe falls below Ours w/ Sim(3) alignment: sharing $\mathbf q_c$ across a submap is what makes it observable.
Without the gate on \cref{eq:reduced_information}, every submap enters $\mathcal U$ and $F_1$ falls below $\mathbf q_c=\mathbf 0$ by $0.4$ and $6.9$ points.
Binned by depth and by camera angular speed (\cref{fig:acc_depth_speed}), \method{} has a lower or comparable median error than Ours w/ Sim(3) variant, most notably in the farthest depth bin.

%% file: tables/tab_model_alignment.tex
\begin{table}[t]
\centering
\captionsetup{font=small}
\caption{Feed-forward model comparison. Mean $F_1$ [\%] $\uparrow$ with Sim(3) alignment (Sim(3)) and ProManO (Ours), at $5$\,cm for indoors and $20$\,cm for outdoors.}
\vspace{-2mm}
\label{tab:model_alignment}
\footnotesize
\setlength{\tabcolsep}{1.6pt}
\begin{tabular*}{\linewidth}{@{}l@{\extracolsep{\fill}}rrrrrrrr@{}}
\toprule
 & \multicolumn{2}{c}{DA3-LARGE} & \multicolumn{2}{c}{DA3-Nested}
 & \multicolumn{2}{c}{Pi3X} & \multicolumn{2}{c}{VGGT-$\Omega$} \\
\cmidrule(lr){2-3} \cmidrule(lr){4-5} \cmidrule(lr){6-7} \cmidrule(lr){8-9}
Dataset & Sim(3) & Ours & Sim(3) & Ours & Sim(3) & Ours & Sim(3) & Ours \\
\midrule
EuRoC & 43.32 & 49.60 & 49.87 & 53.99 & 49.56 & 52.67 & 46.32 & 48.53 \\
CID & 33.33 & 37.23 & 40.49 & 45.24 & 40.29 & 44.11 & 36.65 & 40.99 \\
Oxford & 48.52 & 57.83 & 50.97 & 60.74 & 59.63 & 64.32 & 58.69 & 62.97 \\
KITTI & 22.32 & 28.68 & 24.85 & 29.09 & 25.11 & 29.69 & 24.77& 28.85 \\
\bottomrule
\end{tabular*}
\end{table}

%% file: conclusion.tex
\section{Conclusion}
We showed that, once VIO calibrates the cameras, the transformations compatible with the calibrated intrinsics form a ten-dimensional manifold: Sim(3) plus the three degrees of freedom of the plane at infinity.
\method{} estimates this correction with along-ray anchor uncertainty, a certified similarity synchronization, and an observability gate, and achieves the highest mean $F_1$ on four indoor and outdoor benchmarks.
It assumes an approximately stationary camera-centric bias within each submap, which dynamic objects or large viewpoint changes may violate, and it is sensitive to frontend drift through the VIO anchors.